\documentclass[runningheads]{llncs}
\usepackage[dvipsnames]{xcolor}
\usepackage{url}
\usepackage{graphicx}
\usepackage{tikz}
\usepackage{comment}
\usepackage[accsupp]{axessibility}  % Improves PDF readability for those with disabilities.
\usepackage[bottom]{footmisc}
\usepackage{times}
\usepackage{floatrow}
\usepackage{caption}
\usepackage[toc,page,header]{appendix}
\usepackage{minitoc}
\usepackage[colorlinks, citecolor=green]{hyperref}
\usepackage[utf8]{inputenc}
\usepackage{amsmath,bm, amssymb, amsfonts} % \usepackage is a command that allows you to add functionality to your LaTeX code
\usepackage{dsfont}
\usepackage{multirow}
\usepackage{stmaryrd}
\usepackage{booktabs}
\usepackage{nicefrac}
\usepackage{relsize}
\usepackage{mathabx}
\usepackage{subcaption}
\usepackage{xspace}
\usepackage{array}
\usepackage{mathtools}
\usepackage{tikz}
\usepackage{array}
\newcolumntype{H}{>{\setbox0=\hbox\bgroup}c<{\egroup}@{}}
\usepackage{xspace}
\makeatletter
\DeclareRobustCommand\onedot{\futurelet\@let@token\@onedot}
\def\@onedot{\ifx\@let@token.\else.\null\fi\xspace}
\def\eg{\emph{e.g}\onedot} 
\def\ie{\emph{i.e}\onedot}

\def\vs{\emph{vs}\onedot}

\def\etal{\emph{et al}\onedot}
\makeatother

\DeclareMathOperator*{\argmin}{arg\,min}
\newcommand{\blue}[1]{#1}

\DeclarePairedDelimiter\floor{\lfloor}{\rfloor}
\newcommand{\citep}[1]{\cite{#1}}

\newcommand{\ours}{PoseGPT\xspace}

\newcommand{\causal}{causal\xspace}
\newfloatcommand{capbtabbox}{table}[][\FBwidth]

\usepackage{pgfplots, pgfplotstable}
\pgfplotsset{compat=newest}
\usepgfplotslibrary{fillbetween}

\newcommand{\oxf}{MidnightBlue}

\newcommand{\paris}{RedViolet}

\newcommand{\valc}{PineGreen}

\newcommand{\ourscolor}{Maroon}
\newcommand{\howcolor}{Black}

\newcommand{\oxfmark}{o}
\newcommand{\parmark}{triangle}
\newcommand{\valmark}{square}

\newcommand{\bone}{CornflowerBlue}
\newcommand{\btwo}{MidnightBlue}
\newcommand{\bthree}{TealBlue}
\newcommand{\bfour}{Blue}

\tikzset{every mark/.append style={solid}}
\pgfplotsset{
	grid=both, width=\linewidth, try min ticks=5,
    legend cell align=left, 
    legend style={fill opacity=0.8},
	ylabel near ticks,
    xlabel near ticks,
    every tick label/.append style={font=\footnotesize},
}

\pgfplotsset{
    oxmed/.style={thick, color=\oxf, mark=o},
    oxhard/.style={thick, dashed, color=\oxf, mark=o},
    pamed/.style={thick, color=\paris, mark=star}, 
    pahard/.style={thick, dashed, color=\paris, mark=star},
    val/.style={thick, color=\valc, mark=x},
    oursoxf/.style={thick, color=\ourscolor, mark=\oxfmark},
    howoxf/.style={thick, dashed, color=\howcolor, mark=\oxfmark},
    ourspar/.style={thick, color=\ourscolor, mark=\parmark},
    howpar/.style={thick, dashed, color=\howcolor, mark=\parmark},
    oursval/.style={thick, color=\ourscolor, mark=\valmark},
    howval/.style={thick, dashed, color=\howcolor, mark=\valmark},
    numean/.style={thick, color=\ourscolor, mark=none},
    numin/.style={thick, color=gray, mark=none},
    wid/.style={thick, color=\ourscolor, mark=o, mark size=0.5pt},
    woid/.style={thick, densely dashdotted, color=RedOrange, mark=o, mark size=0.5pt},
    d2k1/.style={thick, color=CornflowerBlue, mark=\oxfmark},
    d2k2/.style={thick, color=MidnightBlue, mark=\oxfmark},
    d2k4/.style={thick, color=TealBlue, mark=\oxfmark},
    d2k8/.style={thick, color=Blue, mark=\oxfmark},
    d4c256/.style={thick, color=\ourscolor, mark=\oxfmark},
    d8c256/.style={thick, color=Green, mark=\oxfmark},
    b1c/.style={thick, color=\bone, mark=o},
    b2c/.style={thick, color=\btwo, mark=x},
    b3c/.style={thick, color=\bthree, mark=*},
    b4c/.style={thick, color=\bfour, mark=o},
    r1c/.style={thick, color=Maroon, mark=o},
    r2c/.style={thick, color=WildStrawberry, mark=o},
    r3c/.style={thick, color=RedViolet, mark=o},
    y1c/.style={thick, color=Goldenrod, mark=o},
    y2c/.style={thick, color=Apricot, mark=o},
    y3c/.style={thick, color=BurntOrange, mark=o},
    tt3/.style={thick, color=Maroon, mark=o},
    tt4/.style={thick, color=BurntOrange, mark=o},
    tt1/.style={thick, color=Maroon, mark=star},
    tt2/.style={thick, color=BurntOrange, mark=star},
    ttl1/.style={thick, color=Maroon},
    ttl2/.style={thick, color=BurntOrange},
    ttl3/.style={thick, mark=star,mark size=3pt},
    ttl4/.style={thick, mark=o},
}

\begin{document}
\pagestyle{headings}
\mainmatter
\def\ECCVSubNumber{6693}  % Insert your submission number here

\title{\ours: Quantization-based 3D Human Motion Generation and Forecasting}

\titlerunning{\ours: Quantization-based 3D Human Motion Generation and Forecasting} 
\def\thefootnote{*}\footnotetext{Equal contribution. \hspace{0.5cm} Accepted as a conference paper at ECCV'22}
%\authorrunning{T. Lucas, F. Baradel, P. Weinzaepfel, G. Rogez} 
\authorrunning{T. Lucas, et al.} 
%\author{Thomas Lucas\thefootnote{*}\orcidlink{0000-0002-6658-6708} \and Fabien Baradel\thefootnote{*}\orcidlink{0000-0003-4625-1713} \and Philippe Weinzaepfel\orcidlink{0000-0002-4223-3983} \and Gr\'egory Rogez\orcidlink{0000-0002-2275-2129}}
\author{Thomas Lucas\thefootnote{*}\and Fabien Baradel\thefootnote{*} \and Philippe Weinzaepfel \and Gr\'egory Rogez}
\institute{NAVER LABS Europe \\[0.05cm] \url{https://europe.naverlabs.com/research/computer-vision/posegpt}}

\maketitle
\begin{abstract}
We address the problem of action-conditioned generation of human motion sequences.
Existing work falls into two categories: forecast models conditioned on observed past motions, or generative models conditioned on action labels and duration only.
In contrast, we generate motion conditioned on observations of arbitrary length, including none.
To solve this generalized problem, we propose \ours, an auto-regressive transformer-based approach which internally compresses human motion into quantized latent sequences.
An auto-encoder first maps human motion to latent index sequences in a discrete space, and vice-versa.
Inspired by the Generative Pretrained Transformer (GPT), we propose to train a GPT-like model for next-index prediction in that space; this allows \ours to output distributions on possible futures, with or without conditioning on past motion.
The discrete and compressed nature of the latent space allows the GPT-like model to focus on long-range signal, as it removes low-level redundancy in the input signal.
Predicting discrete indices also alleviates the common pitfall of predicting averaged poses, a typical failure case when regressing continuous values, as the average of discrete targets is not a target itself.
Our experimental results show that our proposed approach achieves state-of-the-art results on HumanAct12, a standard but small scale dataset, as well as on BABEL, a recent large scale MoCap dataset, and on GRAB, a human-object interactions dataset.
\end{abstract}

\begin{figure}[bht]
\centering
\includegraphics[width=1\linewidth]{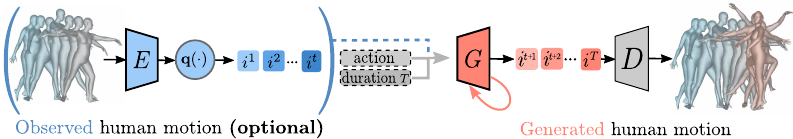} 
\caption{
\textbf{Method Overview.}
\ours generates a human motion sequence, conditioned on an action label, a duration $T$, and optionally on an observed past human motion.
A GPT-like~\citep{Radford2018ImprovingLU} model $G$ sequentially predicts discrete latent indices, which are decoded using a decoder $D$ into a generated human motion.
When conditioning also on past human motion, the input human motion is encoded with $E$ and quantized using $q(.)$ into the discrete latent space.
}
\label{fig:sampling}
\end{figure}

\section{Introduction}
Generating realistic and controllable human motion is still an open research question despite decades of efforts in this domain \cite{Badler1975thesis,Badler1993}.
In this work, we tackle the task of action-conditioned generation of realistic human motion sequences of varying length, with or without observation of past motion.
Most of the effort in human motion synthesis has been focused on future motion prediction, typically conditioned on a sequence of past frames~\citep{Habibie2017ARV,BarsoumCVPRW2018,aksan2019structured,yuan2020dlow,zhang2021we}; however, this requirement is a limiting constraint.
In particular applications to virtual reality or character control \citep{Holden2017PhasefunctionedNN,Starke2019NeuralSM} \blue{ideally should not require real world observations}.
And indeed, recent works \cite{actor,chuan2020action2motion} have shown that deep models can handle the highly multi-modal nature of human motion sequences, without conditioning on the past to narrow it down. %
Nevertheless, many possible applications of human motion modeling do require conditioning.
In particular, vision based human-robot interactions may require robots to observe humans and predict likely future movements to successfully avoid them or interact with them.
Therefore, we propose a class of models flexible enough to approach the more general problem of motion generation conditioned on observations of \emph{arbitrary length}, including none.

Auto-regressive generative models~\citep{oord2016pixel,van2016conditional} are natural candidates to handle this task.
By factorizing distributions over the time dimension, they can be conditioned on past sequences of arbitrary length.
However when applied to human motion sequences their potential is limited in at least two ways by the nature of the data. 
First, they are costly and inefficient to train on data captured at high frame rates, \eg $30$ frames per second (fps), in particular when using state-of-the-art transformer architectures. 
Second, long-term future is highly multi-modal; in a continuous target space this leads to average unrealistic predictions and, in turn, to error drift when sampling from auto-regressive models. 
Indeed, related previous works that have proposed auto-regressive approaches (based on LSTMs \cite{Fragkiadaki2015RecurrentNM} and GRUs \cite{Martinez_2017_CVPR}), have shown that they are subject to error drift and prone to regress unrealistic average poses.

Therefore, we propose to compress human motion into a space that is lower dimensional and  \emph{discrete}, to reduce input redundancy.
This allows training an auto-regressive model using discrete targets rather than to regress in a continuous space, such that the average of targets is not a valid output itself.
We propose an auto-encoder transformer-based network which maps the human motion to a low dimensional space, discretized using a quantization bottleneck~\citep{vqvae}, and vice versa.
Importantly, we ensure that the causal structure of the time dimension is kept in the latent representations such that it \emph{respects the arrow of the time} (i.e. only the past influences the present).
To do so we rely on causal attention in the encoder.
This is crucial to enable conditioning of our model on observed past motions of arbitrary length, unlike in~\cite{actor}. %

Then, we employ an auto-regressive GPT-like model to capture human motion directly in the learned discrete space.
Transformer models have become the de-facto architecture for language tasks \citep{vaswani2017attention,Radford2018ImprovingLU,Radford2019LanguageMA} and are increasingly adopted in computer vision \citep{chen2020generative,Dosovitskiy2021an}.
This requires adaptations to deal with continuous and locally redundant data, which is not well suited to the quadratic computational cost induced by the lack of inductive prior in transformers.
The input data used in this work falls into this category: we employ parametric 3D models \cite{smplx,smpl2015} which represent human motion as a sequence of human 3D meshes, a continuous, high-dimensional and redundant representation.
Our proposed discretization of the human motion alleviates the need for the auto-regressive model to capture low-level signal and enables it to concentrate on long-range relations.
Indeed, while the space of human body model parameters \citep{smplx} is high-dimensional and sparse -- random samples are unlikely to be realistic -- the quantization step concentrates useful regions into a finite set of points.
In particular, random sequences in that space produce locally realistic sequences that lack temporal coherence.
The GPT-like component of our method, called \ours, is trained to predict \blue{a distribution over} the next index in the discrete space. This allows \blue{probabilistic modeling of}
possible futures, with or without conditioning on past motion.

Motion capture (MoCap) datasets with action labels are costly to create \citep{h36m_pami,cmu_mocap}.
We have been able to learn models from several orders of magnitude more data than prior art \citep{actor,chuan2020action2motion}, owing to the recent availability of the BABEL~\cite{babel} dataset, and also relying on the smaller HumanAct12 for fair comparison with previous works.
In addition, we propose an evaluation protocol which we believe aggregates the best practices from prior art~\cite{actor} and from the generative image modeling literature ~\cite{naeem2020reliable,barratt2018note,shmelkov2018good,adeLucas}. 
It is based on three principles; first, sample quality is evaluated using metrics based on classifiers, inspired from the GAN literature. 
Second, we strive to account for over-fitting together with sample quality.
Indeed, sample quality metrics typically compare synthetic data to train data, without employing a validation set, which rewards over-fitting. While this is harmless when working with plentiful and complex data that deep models are unlikely to over-fit, we show that is not the case with small human motion datasets such as HumanAct12.
Finally, we report likelihood based metrics to evaluate mode coverage. Indeed, while it is notoriously difficult to measure diversity from samples alone~\cite{naeem2020reliable,shmelkov2018good}, that is in principle not the case for models that allow likelihood computations on test data. Using these principles we show that our proposed approach outperforms existing ones while being more flexible. \\ % 

\section{Related work}
\noindent\textbf{Human motion forecasting. }
Predicting future human poses given a past motion is a topic of interest in human motion analysis \cite{Badler1975thesis,Badler1993}.
The first successful methods were based on statistical models \cite{bowden2000learning,galata2001learning}, with most recent work relying on deep learning based methods \cite{goodfellow2014generative,kingma2013auto}.
In particular image generation methods such as GANs \cite{goodfellow2014generative} and VAEs \cite{kingma2013auto} have been extended to human motion forecasting \cite{gupta2018social,lin2018human,BarsoumCVPRW2018}.
In DLow~\cite{yuan2020dlow}, a pretrained model is employed to enforce diversity in the predicted motions.
In Cao \etal~\cite{cao2020long}, the scene context is also taken into account to predict future human motion.
However they both show limitations when it comes to predicting long-term future horizons; in particular they tend to  predict average poses, which is a known issue for methods trained by predicting continuous values \cite{ghosh2017learning}.
In contrast, we propose a method able to predict future motion without error drift by quantizing \blue{motions}.

\noindent\textbf{Human motion synthesis. }
The task of human motion synthesis, given a class query,
was first tackled with a focus on simple and cyclic human actions such as walking \cite{urtasun2007modeling,taylor2006modeling}.
More recently, a lot of focus has been devoted to human pose and motion generation conditioned on a rich query representation such as a short textual description \blue{\cite{lin2018human,ahn2018text2action,companim,lang2pose,animlang,posescript}} or an audio representation such as music \cite{li2021learn,lee2019dancing}. \blue{Class labels can be seen as a coarse case of textual descriptions; they bring less information about the motion than detailed descriptions, but are simpler to acquire and use.}
A few recent propositions have tackled 3D human motion generation given action classes \cite{actor,chuan2020action2motion,humor}, and in particular ACTOR \cite{actor} shows impressive results at generating human motion for non-periodic actions.
However only small scale-datasets were available at the time  \cite{chuan2020action2motion,zou2020detailed,zou2020polarization}.
The generated human motions are always front view, and the trajectories in the training data lack diversity. \blue{In \cite{humor}, action sequences are modeled by conditioning predictions at each time frame on the last; this performs well for short sequences but does not allow conditioning on observations of arbitrary length.}
To go beyond these limitations, we develop a method \blue{trained} % that works 
on large-scale datasets, with long-tailed class distributions such as the recently released BABEL \cite{babel}. Our method can optionally be conditioned on past observations of arbitrary length, and obtains state-of-the-art performance. 
\blue{Most similar to ours, the concurrent work in~\citep{bailando} also relies on a quantization step and a GPT model for successfully learning dance motion conditioned on music.} 

\noindent\textbf{Pose representation. }
Human body representations are often expressed as skeleton representations, where a known kinematic structure is available.
Most work in human modeling, ranging from human pose estimation \cite{zheng2020deep,rogez2019lcr,weinzaepfel2020dope,agarwal2005recovering} to human pose modeling \cite{herda2000skeleton,ghosh2017learning}, have used this type of representations for a while.
However recent works are moving toward 3D body shape models \cite{smpl2015,smplx,baradel2021leveraging,kocabas2020vibe} which are more realistic and enable more powerful applications such as augmented and virtual reality.
Representing the 3D human body, and in particular the pose, is not straightforward.
One can express a human pose as a set of 3D joint locations in the Euclidean space or as a set of bone angles encoding the rotations necessary to obtain the pose.
However the lack of continuity in the space of rotation representations is a commonly observed issue \cite{bregier2021deep,zhou2019continuity} for deep learning methods.
There has not been convergence towards a unified human pose representation format so far.
In this work, we do not explicitly enforce any human pose representation but rather
 propose a model that can learn to embed and quantize any representation to a discrete latent space learned by the model.

\noindent\textbf{Generative modeling.}
Deep generative models can be broadly classified in two categories: maximum-likelihood based models, trained to maximize the likelihood of generating training data, and adversarial models \cite{goodfellow2014generative} trained to maximize the quality of generated images as evaluated by a discriminator model. 
In the maximum-likelihood based literature, which is most relevant to our work, there are two dominant paradigms to handle the highly multi-modal nature of perceptual data. The first family is that of variational auto-encoders (VAEs)~\cite{kingma2013auto,vae2}, which relies on an encoder, and the second that of autoregressive models~\cite{oord2016pixel,chen2018pixelsnail} which relies on the chain rule decomposition of high-dimensional data. Both paradigms are leveraged in our work: in the first stage of our approach we adapt a flavour of auto-encoders called VQVAEs \cite{van2017neural}, which uses quantized latent variables, to our problem. In the second stage, we train a transformer based auto-regressive model to sequentially predict discrete latent sequences.  
Similar recipes have been applied to high-resolution image generation in \cite{msp,razavi2019generating,esser2021taming}, to video prediction \cite{weissenborn2020scaling,walker2021predicting} and to speech modeling in \cite{chorowski2019unsupervised}.
Note that while GANs generally display an impressive aptitude to generate high quality samples~\cite{brock2018large}, they are not well suited to the task of human future pose/motion prediction. Indeed, they suffer from mode-collapse~\cite{shmelkov2018good}, \ie, the inability to cover the full variability of the training data. This ability is critical for example for applications such as human-robot interactions where likely modes of the distribution of possible futures must not be ignored. Thus, this class of models is not a good candidate on its own.

\section{The \ours model}
\label{sec:method}

In this section we describe \ours,\footnote{See https://github.com/naver/PoseGPT for implementation details} our proposed approach for  generative modeling of human pose sequences.
First, we present how we compress human motion to a discrete space, and reconstruct motion from it (Section \ref{sec:method:ae}).
Second, we introduce a GPT-like model trained for next-index \blue{probabilistic} prediction in that space (Section \ref{sec:method:gpt}).

\newcommand{\action}{{\mathbf{a}}}
\newcommand{\quantizedcode}{z_{\mathbf{q}}}
\newcommand{\codebookdim}{n_z}
\newcommand{\RR}{\mathbb{R}}
\newcommand{\decoder}{D}
\newcommand{\encoder}{E}
\newcommand{\gpt}{G}
\newcommand{\hz}{\hat{\bm{z}}}
\newcommand{\quantize}{\mathbf{q}}
\newcommand{\codebook}{\mathcal{Z}}
\newcommand{\Attn}{\operatorname{Attn}}
\newcommand{\softmax}{\operatorname{softmax}}

\subsection{Learning a discrete latent space representation}
\label{sec:method:ae}

\begin{figure}[t!]
\centering
\includegraphics[width=0.98\linewidth]{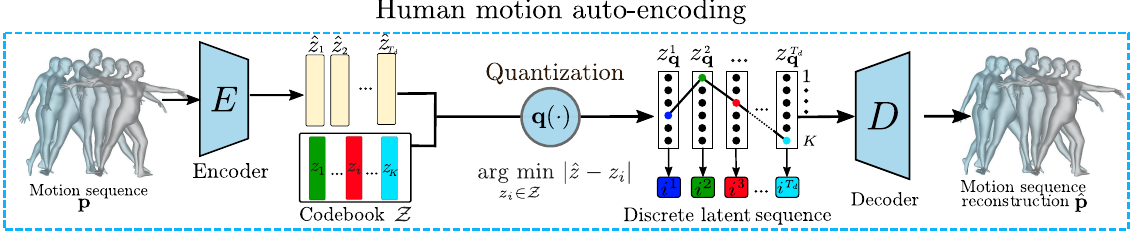}
\caption{\label{fig:training_ae}\textbf{Discrete latent representation for human motion.}
The encoder $\encoder$ maps a human motion $\bm{p}$ to a latent representation $\hat{z}$ which is then quantized using a codebook $\codebook$.
The decoder $\decoder$ reconstructs the human motion $\hat{p}$ from the quantized latent sequence $\quantizedcode$.
}
\end{figure}

\noindent Human actions defined by body-motions can be characterized by the rotations of body parts, disentangled from the body shape.
This allows the generation of motions with actors of different morphology.
For this, we rely on parametric differential body models -- SMPL~\citep{smpl2015}
and SMPL-X~\citep{smplx} --
which disentangle body parts rotations from body shape;
a human motion $\bm{p}$ of length $T$ is represented as a sequence of body poses and translations of the root joints: $\bm{p} = \{(\theta_1, \delta_1), \hdots, (\theta_T, \delta_T)\}$ where $\theta$ and $\delta$ represent the body pose and the translation respectively. 
We use an encoder $\encoder$ and a quantization operator $\quantize$ to encode pose sequences and a decoder $\decoder$ to reconstruct  $\hat{\bm{p}} = \decoder(\quantize(\encoder(\bm{p})).$ 
We use \causal attention mechanisms to maintain a temporally coherent latent space and neural discrete representation learning \cite{vqvae} for quantization.
An overview of the training procedure is shown in Figure \ref{fig:training_ae}.
\\

\begin{figure}[t]
\centering
\includegraphics[width=0.98\linewidth]{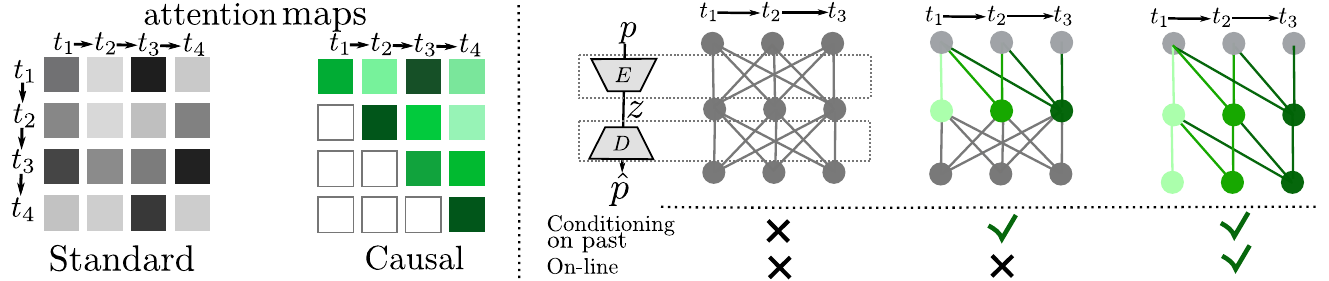}
\caption{\label{fig:causal}\textbf{Conditioning on past with \causal attention.} Masking attention maps in the encoder leads to models that can be conditioned on past observations. Masking the attention maps in the decoder as well allows models that can make on-line predictions.}
\end{figure}

\noindent\textbf{Causal latent space.} 
The encoder first represents human motion sequences as a latent sequence representation $\hz = \{\hat{z}^1, \hdots, \hat{z}^{T_d}\} = \encoder(\bm{p})$ where $T_d \le T$ is the temporal dimension of the latent sequence.
By default, we require that our latent representation respects the arrow of time, \ie, that for any $t \le T_d$, 
$\{\hat{z}_1, \hdots, \hat{z}_t\}$ depends only on $\{p_1, \hdots, p_{\floor*{t \cdot T/T_d}} \}$; such as illustrated in Figure~\ref{fig:causal}. 
For this, we rely on transformers with causal attention; it avoids any inductive prior besides causality, by modeling interactions between all inputs using self-attention~\citep{vaswani2017attention}, modified to respect the arrow of time.
Intermediate representations are mapped using three feature-wise linear projections, into query $Q \in \RR^{N\times d_k}$, key $K \in \RR^{N\times d_k}$
and value $V \in \RR^{N\times d_v}$; in addition, a causal mask is defined as $C_{i,j} = -\infty \cdot \llbracket i > j \rrbracket + \llbracket i \le j \rrbracket$, and the output is computed as:
\begin{equation}
	\Attn(Q,K,V) = \softmax\left(\frac{QK^{\top}\cdot C}{\sqrt{d_k}}\right)V \in
  \RR^{N\times d_v}.
\end{equation}
The causal mask ensures that all entries below the diagonal of the attention matrix do not contribute to the final output and thus that the arrow of time is respected. This is crucial to allow conditioning on past observations when sampling from the model: if latent variables depend on the full sequence, they are impossible to compute from past observations alone. \\

\noindent\textbf{Quantizing the latent space.} 
To build an efficient latent representation of human motion sequences, we then rely on a discrete codebook of learned temporal representations;
more precisely a latent space sequence $\hz \in \RR^{T_d \times \codebookdim}$ is mapped to a sequence of codebook entries $\quantizedcode \in \mathcal{Z}^{T_d}$, where $\mathcal{Z}$ is a set of $C$ codes of dimension $n_z$.
Equivalently, this can be summarized as a sequence of $T_d$ indices corresponding to the code entries in the codebook.
A given sequence $\bm{p}$ is approximately reconstructed by $\hat{\bm{p}}=\decoder(\quantizedcode)$ where $\quantizedcode$ is obtained by encoding $ \hat{z} = \encoder(x) \in \RR^{T_d \times \codebookdim}$ and mapping each temporal element of this tensor with $\quantize(\cdot)$ to its closest codebook entry $z_k$:
\begin{eqnarray}
	\quantizedcode &=& \quantize(\hat{z}) \coloneqq
  \left(\argmin_{z_k \in \codebook} \Vert \hat{z}_{t} - z_k \Vert\right)
  \in \RR^{T_d \times \codebookdim}\\
	\hat{\bm{p}} &=& \decoder(\quantizedcode ) = \decoder\left(
    \quantize(\encoder(\bm{p})) \right).
	\label{eq:vqrec}
\end{eqnarray}

Equation~\eqref{eq:vqrec} is non differentiable; the standard way to backpropagate through it is to rely on the straight-through gradient estimator,
which during the backward pass simply approximates the quantization step as an identity function by copying the gradients from the decoder to the encoder \cite{bengio2013estimating}. Thus the encoder, decoder and codebook can be trained by optimizing: 
\begin{equation}
\mathcal{L}_{\text{VQ}}(\encoder, \decoder, \codebook) = \Vert \bm{p} - \hat{\bm{p}} \Vert^2 + \Vert \text{sg}[\encoder(\bm{p})] - \quantizedcode \Vert_2^2 + \beta \Vert \text{sg}[\quantizedcode] - \encoder(\bm{p}) \Vert_2^2,
\end{equation}
with $\text{sg}[\cdot]$ the stop-gradient operator. The term $\Vert \text{sg}[\quantizedcode] - \encoder(\bm{p}) \Vert_2^2$, dubbed the ``commitment loss'' \cite{vqvae}, has been shown necessary to stable training.
\\

\noindent\textbf{Product quantization.} 
To increase the flexibility of the discrete representations learned by the encoder $\encoder$, we propose using product quantization~\cite{pq}: each element $\hz_i \in \RR^{\codebookdim}$ in the sequence of latent representation is cut into $K$ chunks $(\hz_i^1, \hdots, \hz_i^K) \in \RR^{\codebookdim/K \times K}$, and each chunk is discretized separately using $K$ different codebooks $\{\codebook_1, \hdots, \codebook_K\}.$ The size of the discrete space learned increases exponentially with $K$, for a total of $C^{T_d \cdot K}$ combinations. We empirically validate the utility of using product quantization in our experiments. Instead of one index target per time step, product quantization produces $K$ targets. To capture relations between them, we propose a prediction head that models the $K$ factors sequentially rather than in parallel, called `auto-regressive' head and evaluated in Section \ref{sec:method:gpt}; see the supplementary material for more details. 
\\

\begin{figure}[t!]
\centering
\includegraphics[width=0.98\linewidth]{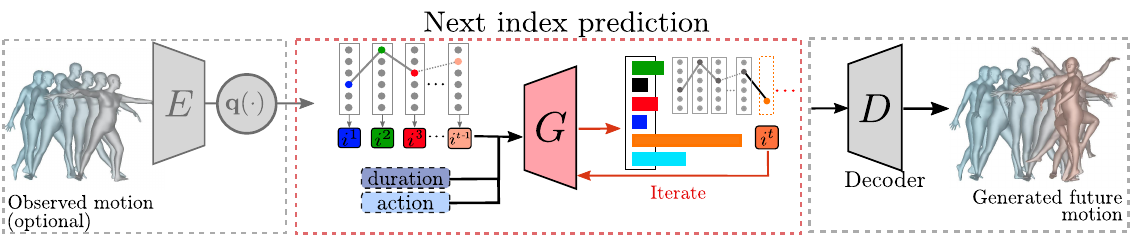}
\caption{\label{fig:training_gpt}\textbf{Future motion prediction.}
In the discrete latent space, an auto-regressive transformer model $\gpt$ predicts the next latent index given previous ones.
We condition on a human action label, a sequence duration and optionally on an observed motion.}
\end{figure}

\subsection{Learning a density model in the discrete latent space}
\label{sec:method:gpt}
The latent representation $\quantizedcode = \quantize(\encoder(\bm{p})) \in \RR^{T_d \times \codebookdim}$ produced by composing the encoder $\encoder$ and the quantization operator $\quantize(\cdot)$ can be represented as the sequence of codebook indices of the encodings,  
$\bm{i} \in \{0, \dots, \vert \codebook \vert - 1 \}^{T_d}$, by replacing each code by its index in the codebook $\codebook$, \ie,
  $i_{t} = k \text{ such that } \left( \quantizedcode \right)_{t} = z_k.$ Indices of $\bm{i}$ can be mapped back to the corresponding codebook entries  and decoded to 
  a sequence $\hat{\bm{p}} = \decoder(z_{i_{1}}, \hdots, z_{i_{T_d}})$.\\

\noindent\textbf{Learning to predict next pose index.} 
As a second step to our method, we propose to learn a prior distribution over learned latent code sequences.
A motion sequence $\bm{p}$ of the human action $a$ is encoded into $(i_t)_{1 .. T_d}$. 
We then formulate the problem of latent sequence generation as auto-regressive index prediction; for this we keep the natural temporal ordering, which can be interpreted as time due to the use of causal attention in the encoder.
We train a transformer model \cite{vaswani2017attention} denoted $\gpt$ -- well suited to discrete sequential data -- using maximum-likelihood estimation, similar in spirit to GPT~\citep{Radford2018ImprovingLU}. 

Given $\bm{i_{<j}}$, the action $a$ and the sequence length $T$, the transformer \blue{outputs a softmax distribution} over the next indices, \ie,      $p_{\gpt}(i_j\vert \bm{i_{<j}}, a, T)$, the likelihood of the latent sequence is $p_{\gpt}(\bm{i})=\prod_j p_{\gpt}(\mathbf{i}_j \vert \mathbf{i}_{<j}, a , T)$ and the model is trained to minimize:
\begin{equation}
  \mathcal{L}_{\text{GPT}} = \mathbb{E}_{\bm{i}} \left[ - \sum_j \log p_{G}(i_j \vert \bm{i_{< j}}, a , T) \right].
\end{equation}
An overview of the training procedure is shown in Figure \ref{fig:training_gpt} and, in the supplementary material, we discuss different \blue{input sequence embeddings for processing by the GPT.}

\begin{figure}[t!]
\centering
\includegraphics[width=0.98\linewidth]{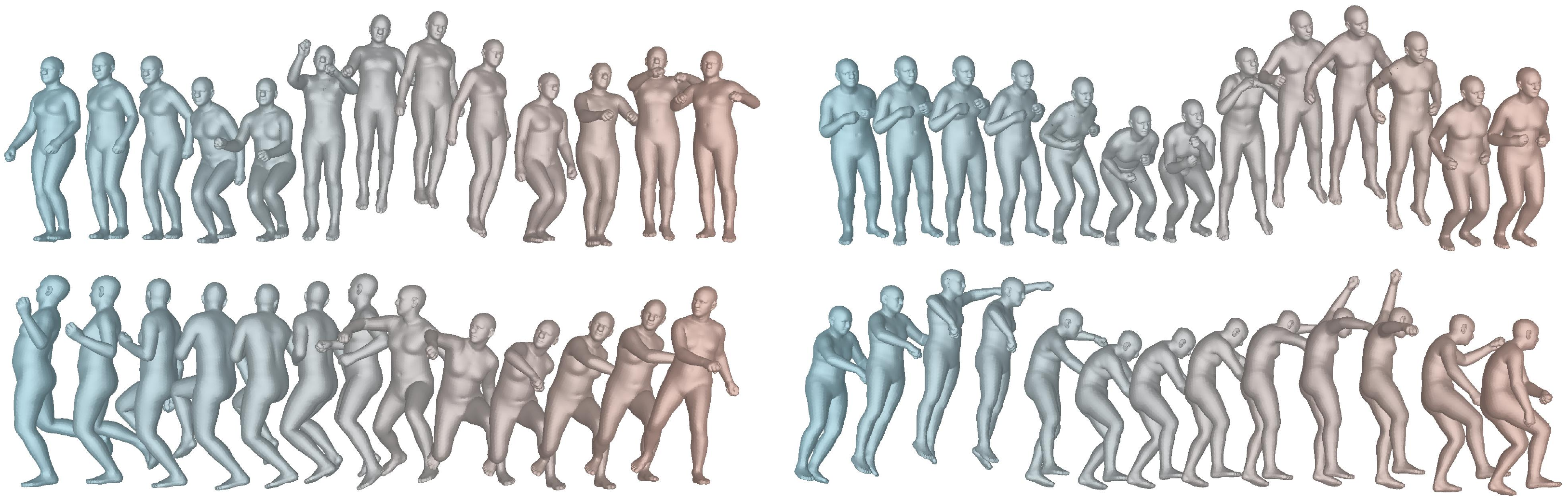} \\[-0.3cm]
\caption{\label{fig:samples_1}\textbf{Samples generated from scratch. }
Samples generated without any observed motion for the action labels `jumping' (top) and for the action `dancing' (bottom).
Note: Times flows from left to right (\ie, the blue texture corresponds to the first frame and the red texture to the last frame).
}
\end{figure}

\noindent\textbf{Sampling human motion.}
Human motion is generated sequentially by sampling from $p(s_i | \bm{s_{<i}}, a, T)$ to obtain a sequence of pose indices $\tilde{\bm{z}}$ given an action and sequence length, and decoding it into a sequence of pose  $\tilde{\bm{p}}= \decoder(\tilde{\bm{z}})$ (see Figure \ref{fig:samples_1} for samples).

\section{Experiments}
\label{sec:expe}

We experiment with two parametric 3D models: SMPL \cite{smpl2015} for comparison to state-of-the-art approaches, and SMPL-X \cite{smplx} to enable control of the face and hands. 
We now present the three datasets considered for evaluation; architectural and implementation details are in \blue{the supplementary material}.

\noindent \textbf{HumanAct12}
allows comparison to prior art \cite{actor,chuan2020action2motion}, but its small size and the absence of train/val/test splits are limiting. It contains $1191$ videos and SMPL pose parameters, 12 action classes and a single action per video.
The poses, automatically optimized from estimated 3D joints, are noisier than annotations from capture environments.

\noindent \textbf{BABEL}
\cite{babel} is a subset of AMASS \cite{AMASS:ICCV:2019}, a large collection of MoCap data captured in controlled environments for high quality annotations. 
It contains  $28$K sequences ($43$ hours of motion in total); sequence length varies from 3 seconds to several minutes and there are $120$ manually annotated human actions in total. 
The action distribution is very long-tailed so we use only the $60$ most common actions as proposed by the authors. 
In short, BABEL is over $40$ times bigger than HumanAct12, has a train/val/test split, no noise in the SMPL parameters and a rich variety of human actions; we believe this makes it a dataset of choice to move forward.

\noindent\textbf{GRAB}
\cite{GRAB:2020} contains whole-body SMPL-X of people grasping objects,
with 11 persons performing 29 motions with 51 different rigid objects, for a total of $1500$ sequences of $8$ seconds on average, with
$7$ persons for training and $2$ for testing.

\begin{figure}[tbh]
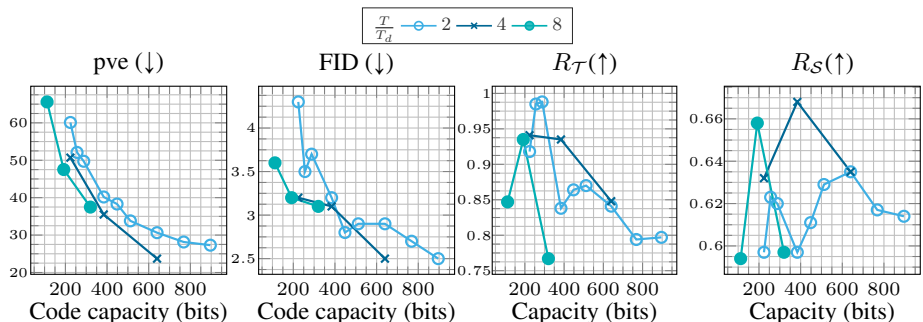

\begin{center}
\begin{subfigure}{.40\linewidth}
\begin{center}
\resizebox{.6\linewidth}{!}
{
    \begin{tikzpicture}
    % \begin{axis}[%
    %   width=.8\linewidth,
    % %   height=6cm, 
    %     legend columns = 5,
    %     hide axis,
    %   ]
    \begin{axis}[%
        hide axis,
        xmin=10, xmax=50,
        ymin=0,ymax=0.4,
        legend columns=-1,
        legend style={draw=white!15!black,legend cell align=left}
        ]
        \addlegendimage{empty legend}
        %\addlegendimage{}
        \addlegendentry{$\frac{T}{T_d}$}
        \addlegendimage{b1c}
        \addlegendentry{$2$}
        \addlegendimage{b2c} 
        \addlegendentry{$4$}
        \addlegendimage{b3c}
        \addlegendentry{$8$}
    \end{axis}
	%\pgfresetboundingbox
	%\path
          %(current axis.south west) -- ++(0.0in,-0.0in)
          %rectangle (current axis.north east) -- ++(0.0in,0in)
    \end{tikzpicture}
    }
    \end{center}
\end{subfigure}
\resizebox{\linewidth}{!}{
    \begin{subfigure}{.24\linewidth}
        \centering
        \begin{tikzpicture}
\begin{axis}[%
  width=1.4\linewidth,
  height=4cm, 
   xlabel={Code capacity (bits)},
   xlabel shift=-0.05in,
  yticklabel style = {font=\tiny},
  xticklabel style = {font=\scriptsize},
  yticklabel shift=-0.03in,
  minor tick num=3,
%   ymin=0.595, ymax=0.645,
    % xmin=50, xmax=700,
 %legend size=small,
  legend pos=north east,
    % legend pos=outer north east,
    % xmode=log,
    title={pve ($\downarrow$)},
            title style={yshift=-5pt},
  ]

\input{plots/pgf_read_table}
    \addplot[b1c]      table[x=D2_kl,  y=D2_pve]   \map; 
    \addplot[b2c]      table[x=D4_C256_kl,  y=D4_C256_pve]   \map; 
    \addplot[b3c]      table[x=D8_C256_kl,  y=D8_C256_pve]   \map; 
\end{axis}
\pgfresetboundingbox
\path
  (current axis.south west) -- ++(0.9in,0.0in)
  rectangle (current axis.north east) -- ++(-0.8in,0.2in);
\end{tikzpicture}
        \label{fig:ablation_nfeat_all_scales_par}
    \end{subfigure}%
    %\hspace*{-0.5cm}
    \begin{subfigure}{.24\linewidth}
        \centering
        \begin{tikzpicture}
\begin{axis}[%
  width=1.4\linewidth,
  height=4cm, 
   xlabel={Code capacity (bits)},
  minor tick num=3,
   xlabel shift=-0.05in,
  %x dir=reverse,
%   ymin=0.595, ymax=0.645,
    % xmin=50, xmax=700,
  legend pos=north east,
  yticklabel style = {font=\tiny},
  yticklabel shift=-0.03in,
  xticklabel style = {font=\scriptsize},
    % legend pos=outer north east,
    % xmode=log,
    title={FID ($\downarrow$)},
            title style={yshift=-5pt},
  ]

\input{plots/pgf_read_table}
    \addplot[b1c]      table[x=D2_kl,  y=D2_FID]   \map; 
    \addplot[b2c]      table[x=D4_C256_kl,  y=D4_C256_FID]   \map; 
    \addplot[b3c]      table[x=D8_C256_kl,  y=D8_C256_FID]   \map; 
\end{axis}
\pgfresetboundingbox
\path
  (current axis.south west) -- ++(0.0in,0.0in)
  rectangle (current axis.north east) -- ++(-0.2in,0.2in);
\end{tikzpicture}
        \label{fig:ablation_nfeat_all_scale_ox}
    \end{subfigure}%
    %\hspace*{-0.5cm}
    \begin{subfigure}{.24\linewidth}
        \centering
        \begin{tikzpicture}
\begin{axis}[%
  width=1.35\linewidth,
  height=4cm, 
   xlabel={Capacity (bits)},
  minor tick num=3,
   xlabel shift=-0.05in,
  legend pos=north east,
  yticklabel style = {font=\tiny},
  xticklabel style = {font=\scriptsize},
  yticklabel shift=-0.04in,
    % legend pos=outer north east,
    % xmode=log,
    title={$R_\mathcal{T}$($\uparrow$)},
            title style={yshift=-5pt},
  ]

\input{plots/pgf_read_table}
    \addplot[b1c]      table[x=D2_kl,  y=D2_ARN]   \map; 
    \addplot[b2c]      table[x=D4_C256_kl,  y=D4_C256_ARN]   \map; 
    \addplot[b3c]      table[x=D8_C256_kl,  y=D8_C256_ARN]   \map; 
    %\node[] at (axis cs: 525,65.6) {1000};
    %\node[] at (axis cs: 810,70) {2000};
    %\node[\ourscolor] at (axis cs: 190,72.8) {1000};
    %\draw[fill, \bt] (axis cs: 400,21) circle (2pt);
    % \draw[fill] (axis cs: 815.4312037,68.875) circle (2pt);
    % \draw[fill,\ourscolor] (axis cs: 364.8590026,72.41) circle (2pt); 
\end{axis}
\pgfresetboundingbox
\path
  (current axis.south west) -- ++(0.5in,0.0in)
  rectangle (current axis.north east) -- ++(-0.1in,0.2in);
\end{tikzpicture}
        \label{fig:ablation_nfeat_all_scale_ox}
    \end{subfigure}%
    %\hspace*{-0.5cm}
    \begin{subfigure}{.24\linewidth}
        \centering
        \begin{tikzpicture}
\begin{axis}[%
  width=1.4\linewidth,
  height=4cm, 
   xlabel={Capacity (bits)},
  minor tick num=3,
   xlabel shift=-0.05in,
%   ymin=0.595, ymax=0.645,
    % xmin=50, xmax=700,
 % legend size=small,
  legend pos=north east,
  yticklabel style = {font=\tiny},
  xticklabel style = {font=\scriptsize},
  yticklabel shift=-0.04in,
    % legend pos=outer north east,
    % xmode=log,
    title={$R_\mathcal{S}$($\uparrow$)},
            title style={yshift=-5pt},
  ]

\input{plots/pgf_read_table}
    \addplot[b1c]      table[x=D2_kl,  y=D2_ASN]   \map; 
    \addplot[b2c]      table[x=D4_C256_kl,  y=D4_C256_ASN]   \map; 
    \addplot[b3c]      table[x=D8_C256_kl,  y=D8_C256_ASN]   \map; 
\end{axis}
\pgfresetboundingbox
\path
  (current axis.south west) -- ++(-0.1in,0.0in)
  rectangle (current axis.north east) -- ++(-0.3in,0.2in);
\end{tikzpicture}
        \label{fig:ablation_nfeat_all_scale_ox}
    \end{subfigure}}
\end{center}
\caption{\label{fig:ae_babel}\small \textbf{Latent space design.} We define models for $T/T_d \in \{2, 4, 8\}$ by varying $K$ and $C$ and present results as a function of the capacity of latent sequence. }
\label{fig:ablation_quant}
\end{figure}

\subsection{Evaluation metrics} 
Generative models can be evaluated through generated data;
a perfect set of samples contains data that is \emph{as realistic} and \emph{as diverse} as real \emph{unseen} test data.
These aspects are not always trivial to quantify, and we now discuss how they are measured in practice.

\noindent \textbf{Sample quality evaluation.} 
The dominant approach~\citep{actor,chuan2020action2motion} to measure sample quality relies on pretrained classifiers. 
In particular the Frechet Inception Distance (FID), which we report, measures a distance between distributions of classifier features obtained from a set of samples $D_\text{samples}$ and real data.
Following \cite{actor}, we also rely on a classifier $\mathcal{T}$ pre-trained on train data and report the ratio between accuracies on sampled and test data:
\begin{equation}
R_{\mathcal{T}}(D_\text{samples}, D_\text{test}) =  \frac{|D_\text{samples}|}{|D_\text{test}|}\cdot\frac{\sum_{x \in D_\text{test}}\text{acc}_{\mathcal{T}}(x)}{\sum_{x \in D_\text{samples}}\text{acc}_{\mathcal{T}}(x)}.
\end{equation}
This metric is not sensitive to diversity -- the model can drop modes as long as the rest is very well classified. 
The ratio normalizes values that otherwise depend on choices orthogonal to sample quality; we refer to the supplementary material for details on the action classifier.

\noindent \textbf{Diversity evaluation.}
First, we evaluate sample diversity by training a classifier $\mathcal{S}$ on \emph{samples} and evaluating it on unseen \emph{test data}, following \cite{shmelkov2018good}. Intuitively, for $\mathcal{S}$ to perform as well as $\mathcal{T}$, samples need to be as diverse and as realistic as real data; we measure it with: 
\begin{equation}
R_{\mathcal{S}}(D_\text{test}) =  \sum_{x \in D_\text{test}}\frac{\text{acc}_{\mathcal{S}}(x)}{\text{acc}_{\mathcal{T}}(x)}.
\end{equation}
This metric is sensitive to diversity as real data modalities not captured by the generator will not be seen by $\mathcal{S}$ and  misclassified, but not by $\mathcal{T}$, which will degrade the ratio.  The pair ($R_\mathcal{S}$, $R_\mathcal{T}$) is best considered together \cite{shmelkov2018good}: if $R_\mathcal{S}$ is close to one, we consider sample quality to be high, and gains in $R_\mathcal{T}$ can be attributed to diversity \cite{naeem2020reliable}. \blue{Note that $\mathcal{S}$ and $\mathcal{T}$ have the same architecture and are trained with the same hyper parameters}. More classically, we also report likelihood based metrics; dropped modes will lead to data points with very low likelihood, so they are sensitive to mode coverage \cite{barratt2018note}.
In particular, we report the test reconstruction error of the auto-encoder using the Per-Vertex Error (pve),
and the test likelihood of the GPT on encoded test sequences. As these metrics do not guarantee realistic samples, we consider them together with classifier based quality metrics. \\

\noindent \textbf{Over-fitting.} Sample quality metrics typically used -- standard FID or classification accuracy~\cite{actor} -- measure differences between train data and generated data, without involving a test set. This does not account for over-fitting and rewards models that perfectly copy train data: on small datasets, all metrics will monotonically improve with model capacity. To remedy this, we keep unseen data on BABEL and compute the FID, $R_{\mathcal{S}}$ ratio and maximum-likelihood based metrics using that test data. Our only metric not sensitive to over-fitting is $R_{\mathcal{T}}$; we rely on the others to detect over-fitting. \\

\begin{table}[!t]
\footnotesize
\centering
    \begin{tabular}{cc}
    \resizebox{0.6\linewidth}{!}{
    \newcolumntype{C}{@{\extracolsep{0.2cm}}c@{\extracolsep{0pt}}}
    \begin{tabular}{cccccCcccCccc}
    \toprule
    \begin{tabular}{c}$K$ \\[-0.2cm] {\tiny (nb. codebooks)  }\end{tabular}
	& pve$\downarrow$ & $R_\mathcal{T}$$\uparrow$ & $R_\mathcal{S}$$\uparrow$ & $\text{FID}$$\downarrow$ &
     pve$\downarrow$ & $R_\mathcal{T}$$\uparrow$ & $R_\mathcal{S}$$\uparrow$ & $\text{FID}$$\downarrow$ &
	 pve$\downarrow$ & $R_\mathcal{T}$$\uparrow$ & $R_\mathcal{S}$$\uparrow$ & $\text{FID}$$\downarrow$ \\
        \midrule 
         & \multicolumn{4}{c@{\hskip 0.4cm}}{$C=128$} & \multicolumn{4}{c@{\hskip 0.4cm}}{$C=256$} & \multicolumn{4}{c@{\hskip 0.4cm}}{$C=512$}\\
        %\midrule
        \cmidrule(lr){1-1} \cmidrule(lr){2-5} \cmidrule(lr){6-9} \cmidrule(lr){10-13}
	% 128
	1   &   60.1        &   \bf{0.91}         &   0.60              &   4.3             &   52.1         &   \bf{0.98}        &   0.61   &   \underline{3.5}              &   49.7        &   \underline{\bf{0.99}}        &   \underline{0.62}                    &  3.7     \\
	2   &   40.2        &   0.84   	          &   0.60              &   3.2             &   38.3         &   0.86             &   0.61                    &   \underline{2.8}               &   33.8        &   \underline{0.87}             &   \underline{\bf{0.63}}   		&  2.9     \\
	4   &   30.6        &   \underline{0.84}  &   \underline{\bf{0.64}}         &   \bf{2.9}        &   28.2         &   0.79             &   \bf{0.62}               &   \bf{2.7}         	       &   27.3        &   0.80             &   0.61        		&  \underline{\bf{2.5}}\\
	8   &   27.2        &   \underline{0.48}   	          &   \underline{0.49}              &   \underline{3.0}             &   23.7         &   0.49             &   0.46                    &   4.4              	       &   26.7        &   0.41             &   0.47        		&  4.3     \\

	\bottomrule
    \end{tabular}} & 
    \resizebox{0.4\linewidth}{!}{
    \newcolumntype{C}{@{\extracolsep{0.2cm}}c@{\extracolsep{0pt}}}
     \renewcommand{\arraystretch}{1.25}
    \begin{tabular}{cccccCccc}
    \toprule
    \begin{tabular}{c}$K$ \\[-0.2cm] {\tiny (nb. codebooks) }\end{tabular}
	& pve$\downarrow$ & $R_\mathcal{T}$$\uparrow$ & $R_\mathcal{S}$$\uparrow$ & $\text{FID}$$\downarrow$ &
    pve$\downarrow$ & $R_\mathcal{T}$$\uparrow$ & $R_\mathcal{S}$$\uparrow$ & $\text{FID}$$\downarrow$ \\
	\midrule 
    & \multicolumn{4}{c@{\hskip 0.4cm}}{$T/T_d=4$} & \multicolumn{4}{c@{\hskip 0.4cm}}{$T/T_d=8$} \\
    %\midrule
\cmidrule(lr){1-1} \cmidrule(lr){2-5} \cmidrule(lr){6-9}

    2 & 50.7 & \bf{0.94} & 0.63      & 3.2       & 47.5 & \bf{0.93} & \bf{0.66} & 3.2      \\
	4 & 35.5 & 0.93      & \bf{0.67} & 3.1       & 37.5 & 0.77      & 0.60      & \bf{3.1} \\
	8 & 28.8 & 0.64      & 0.54      & \bf{2.5}  & 65.6 & 0.85      & 0.59      & 3.6      \\
	\bottomrule
    \end{tabular}}\\
    \noalign{\vskip 0.5mm} 
    {\scriptsize ($T/T_d = 2$)} &{\scriptsize ($C=256$)}
    \end{tabular} \\[-0.3cm]
    \caption{
    \label{tab:ae_babel}
    \small \textbf{Impact of latent space capacity} on BABEL for $T/T_d=2$ (left) and for $C=256$ (right). \textbf{bold} denotes best in column (across $K$); \underline{underlined} denotes best in row (across $C$).}
\end{table}

\subsection{Ablative study of design choices} 
% \begin{figure}[tbh]

\begin{figure}[!b]
\begin{floatrow}
\capbtabbox[5cm]{%
\resizebox{1.0\linewidth}{!}{
  \renewcommand{\arraystretch}{1.1}
 \newcolumntype{C}{@{\extracolsep{0.2cm}}c@{\extracolsep{0pt}}}
  \begin{tabular}{cCcCc}
    \toprule
    \begin{tabular}{c}$K$ \\[-0.2cm] {\tiny (nb. codebooks)  }\end{tabular} &
	{\scriptsize FID $\downarrow$}& %$\downarrow$
	{\scriptsize $R_\mathcal{S} (\uparrow)$}&
	{\scriptsize FID $\downarrow$}&
	{\scriptsize $R_\mathcal{S} (\uparrow)$}\\
	\midrule
	 & \multicolumn{2}{c}{$C=256$} & \multicolumn{2}{c}{$C=512$} \\
     \cmidrule(lr){1-1} \cmidrule(l){2-3} \cmidrule(lr){4-5}

        8  & 0.12         & 93.8   & 0.11         & 93.7     \\
        16 & 0.11         & 94.5   & 0.10         & 94.9     \\
        32 & \bf{0.09}         & \bf{95.1}   & \bf{0.08}    & \bf{95.2}\\
	\bottomrule
    \end{tabular}
    }
}{%
  \caption{\label{tab:ae_humanact}\footnotesize \textbf{Latent space design} on HumanAct12. \textbf{Bold} denotes best value.}%
}
\ffigbox[7cm]{%
\begin{center}
\resizebox{0.4\linewidth}{!}{
%\begin{center}
\resizebox{.6\linewidth}{!}{
    \begin{tikzpicture}
    \begin{axis}[%
        hide axis,
        xmin=10, xmax=50,
        ymin=0,ymax=0.4,
        legend columns=-1,
        legend style={draw=white!15!black,legend cell align=left}
        ]
        \addlegendimage{empty legend}
        %\addlegendimage{}
        \addlegendentry{$C$}
        \addlegendimage{r1c}
        \addlegendentry{$128$}
        \addlegendimage{r2c} 
        \addlegendentry{$256$}
        \addlegendimage{r3c}
      (current axis.south west) -- ++(0.0in,0.0in)
  rectang    \addlegendentry{$512$}
    \end{axis}
    \end{tikzpicture}
    }
    %\end{center}
}\\
\resizebox{0.49\linewidth}{!}{
%\begin{center}
\begin{tikzpicture}
\begin{axis}[%
%   width=.5\linewidth,
  height=5cm, 
   xlabel={Code Capacity (bits)},
  minor tick num=3,
 % x dir=reverse,
%   ymin=0.595, ymax=0.645,
    % xmin=50, xmax=700,
  %legend size=small,
  legend pos=north east,
    % legend pos=outer north east,
    % xmode=log,
    title={Compr. cost of $\quantizedcode$ (bits)},
            title style={yshift=-5pt},
  ]

\input{plots/pgf_read_table}
    \addplot[r1c]      table[x=D2_C128_kl,  y=D2_C128_gbits]   \map; 
    \addplot[r2c]      table[x=D2_C256_kl,  y=D2_C256_gbits]   \map; 
    \addplot[r3c]      table[x=D2_C512_kl,  y=D2_C512_gbits]   \map; 
\end{axis}
\pgfresetboundingbox
\path
  (current axis.south west) -- ++(0.0in,0.0in)
  rectangle (current axis.north east) -- ++(0.0in,0.2in);
\end{tikzpicture}
%\end{center}
}
\resizebox{0.49\linewidth}{!}{
\begin{tikzpicture}
\begin{axis}[%
%   width=.5\linewidth,
  height=5cm, 
   xlabel={Code capacity (bits)},
  minor tick num=3,
%   ymin=0.595, ymax=0.645,
    % xmin=50, xmax=700,
  %legend size=small,
  legend pos=north east,
    % legend pos=outer north east,
    % xmode=log,
    title={Compr. cost (bits per dim)},
            title style={yshift=-5pt},
  ]

\input{plots/pgf_read_table}
    \addplot[r1c]      table[x=D2_C128_kl,  y=D2_C128_gnll]   \map; 
    \addplot[r2c]      table[x=D2_C256_kl,  y=D2_C256_gnll]   \map; 
    \addplot[r3c]      table[x=D2_C512_kl,  y=D2_C512_gnll]   \map; 
\end{axis}
\pgfresetboundingbox
\path
  (current axis.south west) -- ++(0.0in,0.0in)
  rectangle (current axis.north east) -- ++(0.0in,0.0in);
\end{tikzpicture}}
\end{center}
}{%
  \caption{\label{fig:compression}\footnotesize \textbf{Cost of compressing $\quantizedcode$} using the GPT, in bits and bits per dimension. }%
  }
\end{floatrow}
\end{figure}

We now ablate the main design choices made in \ours. The first is the design of the discrete latent space, in particular the quantization bottleneck and its capacity.
The second regards the GPT component, trained for next index prediction; in particular we ablate the choice of input embedding method and prediction head. Finally, we evaluate the impact of using causal attention in the auto-encoder. Note that as there is no test split on HumanAct12; because it is too small to define one of reasonable size without severely degrading performance, we compute the FID using train data on this dataset. \\

\noindent\textbf{Latent sequence space design.}
The main design choice regarding the latent sequence space is the quantization bottleneck. We now study the impact of its capacity, mostly controlled by $T_d$ (latent sequence length), $K$ (nb. of product quantization factor) and $C$ (total number of centroids). 
 More capacity yields lower reconstruction errors at the cost of less compressed representations. In our case, that means more indices to predict for the GPT, which impacts sampling, and we now explore this trade-off.
 
In Table \ref{tab:ae_babel} (left), models trained on BABEL show that as expected, pve goes down monotonously with both $K$ and $C$, but not the $R_\mathcal{S}$ and $R_\mathcal{T}$ ratios, as also shown in Figure~\ref{fig:ae_babel}. Models with $K = 1$ obtain high sample classification accuracy but poor reconstruction on test data and lower $R_\mathcal{S}$; this suggests insufficient capacity to capture the full diversity of the data. On the other hand, models with the most capacity (\eg, $K = 8$) yield sub-par performance. The best trade-offs are achieved with $(K,C) \in \{(2,256), (2,512), (4,128), (4,256)\}$. The table on the right shows that the model can handle decreased temporal resolution. Note that using $K=8$ works better at coarser resolutions, as it compensates for the loss of information. In Table \ref{tab:ae_humanact}, all metrics improve monotonically with $K$ and $C$; this is expected as over-fitting is not factored out by the metrics and the dataset is small enough to over-fit. Finally, in Figure~\ref{fig:compression}, we report the cost of compressing $\quantizedcode$ using the GPT model. We observe that the absolute compression cost in bits (left) increases, \ie, $\quantizedcode$ contains more information, while the cost per dimension decreases: each sequence index is easier to predict individually. 

\begin{figure}[!t]
\begin{floatrow}
\capbtabbox[7.5cm]{%
\resizebox{0.9\linewidth}{!}{
%\begin{center}
\resizebox{1\linewidth}{!}
{
    \begin{tikzpicture}
    % \begin{axis}[%
    %   width=.8\linewidth,
    % %   height=6cm, 
    %     legend columns = 5,
    %     hide axis,
    %   ]
    \begin{axis}[%
        hide axis,
        xmin=10, xmax=50,
        ymin=0,ymax=0.4,
        legend columns=-1,
        legend style={draw=white!15!black,legend cell align=left}
        ]
        \addlegendimage{empty legend}
        %\addlegendimage{}
        %\addlegendentry{}
        %\addlegendimage{y1c}
        \addlegendentry{$t^0$: extra token at $t=0$. \hspace{0.3cm}}
        \addlegendimage{empty legend}
        \addlegendentry{$+$: sum at all $t$. \hspace{0.3cm}}
         \addlegendimage{empty legend}
        \addlegendentry{$::$ concat at all $t$. \hspace{0.3cm}}
        %\addlegendimage{empty legend}
        %\addlegendentry{$\times$: feature missing}
          %\addlegendentry{cat}
        %\addlegendimage{y3c}
        %\addlegendentry{cat,ar}
    %\addlegendentry{\texttt{cat \& ar}}
    \end{axis}
    \end{tikzpicture}
    }
%    \end{center}
}
\resizebox{1.\linewidth}{!}{
 \begin{tabular}{cc@{\hskip 0.25cm}|@{\hskip 0.25cm}cc@{\hskip 0.25cm}|@{\hskip 0.25cm}ccHc@{\hskip 0.25cm}|@{\hskip 0.25cm}lr}
 \toprule 
 \multicolumn{2}{l@{\hskip 0.5cm}}{Input tokens} & \multicolumn{2}{c@{\hskip 0.5cm}}{Prediction head} & \multicolumn{4}{c@{\hskip 0.5cm}}{BABEL}  & \multicolumn{2}{c}{HumanAct12} \\
 \midrule
    $a$ & $T$ & mlp & ar & $R_\mathcal{S}(\uparrow)$ & $R_\mathcal{T}(\uparrow) $ & FID($\downarrow$) & gpt-acc.($\uparrow$)& FID($\downarrow$)   & $R_{\mathcal{T}}(\uparrow)$ \\
    \midrule
        {\small $t^0$} &  $\times$  &  $\times$ & $\times$  & 0.65 & 0.51 & - & 22.4 {\tiny (20.8)} & 0.56 & 0.74\\
        +   &$\times$ &$\times$&$\times$& 0.79 & 0.55 & - & 23.1 {\tiny (16.3)} & 0.19 & 0.91 \\
        +   & +  &$\times$&$\times$& 0.79 & 0.57 & -  & 23.9 {\tiny (16.2)} & \bf{0.10} & \bf{0.94} \\
     	::  & :: &$\times$&$\times$& \bf{0.86} & \bf{0.61} & 2.8 & \bf{24.9} {\tiny (22.9)} & 0.22  & 0.86 \\
	\midrule
	+    & +   & \checkmark &$\times$& - & - & - & - & \bf{0.09} & \bf{95.1} \\
	\midrule
	::   & ::  & \checkmark &$\times$& 0.86 & 0.61 & 2.8 & 25.1{\tiny (22.7)} & - & - \\
	::   & ::  & \checkmark & \checkmark & \bf{0.98} & \bf{0.64} & 2.8 & \bf{25.9{\tiny (22.6)}} & - & - \\
	\bottomrule
    \end{tabular}}
}{%
  \caption{\label{tab:abl_input}\footnotesize \textbf{GPT design} on BABEL and HumanAct12. \textbf{Bold} text denotes best value. ar denotes auto-regressive.}%
}
\ffigbox[4.5cm]{%
\begin{center}
\resizebox{0.7\linewidth}{!}{
%\begin{center}
\resizebox{.6\linewidth}{!}{
    \begin{tikzpicture}
    % \begin{axis}[%
    %   width=.8\linewidth,
    % %   height=6cm, 
    %     legend columns = 5,
    %     hide axis,
    %   ]
    \begin{axis}[%
        hide axis,
        xmin=10, xmax=50,
        ymin=0,ymax=0.4,
        legend columns=-1,
        legend style={draw=white!15!black,legend cell align=left}
        ]
        \addlegendimage{empty legend}
        %\addlegendimage{}
        \addlegendentry{}
        \addlegendimage{y1c}
        \addlegendentry{\texttt{sum}}
        %\addlegendentry{sum}
        \addlegendimage{y2c} 
        \addlegendentry{\texttt{cat}}
          %\addlegendentry{cat}
        \addlegendimage{y3c}
        %\addlegendentry{cat,ar}
    \addlegendentry{\texttt{cat \& ar}}
    \end{axis}
    \end{tikzpicture}
    }
    %\end{center}
}\\
%\hspace*{-0.5cm}
\resizebox{0.9\linewidth}{!}{
\begin{tikzpicture}
\begin{axis}[%
%   width=.5\linewidth,
  height=4.1cm, 
   xlabel={{\tiny Training iterations}},
   xlabel shift=-1.1ex,
  minor tick num=3,
%   ymin=0.595, ymax=0.645,
    % xmin=50, xmax=700,
  %legend size=small,
  legend pos=north east,
    % legend pos=outer north east,
    % xmode=log,
    %tick label style={font=tiny}
    yticklabel style = {font=\tiny,xshift=0.5ex},
    xticklabel style = {font=\tiny,yshift=0.5ex},
    title={{\tiny Val. accuracy}},
            title style={yshift=-8pt},
  ]
\pgfplotstableread{
iter    sum     cat    mlp_ar
100     6.5     14.3   14.5
200     12.7    17.7   18.1
300     16.7    19.8   20.1
400     18.0    20.8   21.2
500     19.9    22.1   22.5
600     21.1    23.0   23.5
}{\map}

    \addplot[y1c]      table[x=iter,  y=sum]   \map; 
    \addplot[y2c]      table[x=iter,  y=cat]   \map; 
    \addplot[y3c]      table[x=iter,  y=mlp_ar]   \map; 
\end{axis}
\pgfresetboundingbox
\path
  (current axis.south west) -- ++(0.0in,0.1in)
  rectangle (current axis.north east) -- ++(0.0in,0.1in);
\end{tikzpicture}}
\end{center}
}{%
  \caption{\label{fig:sumcat}\footnotesize \textbf{Index pred. accuracy} using concatenation \vs summation.}%
}
\end{floatrow}
\end{figure}

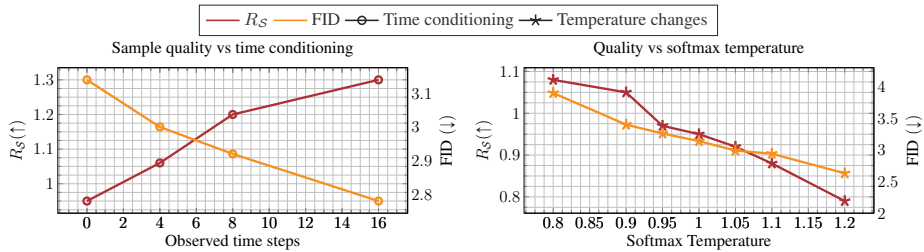
\begin{figure}[!b]
\begin{center}
%\hspace*{-0.4cm}
\resizebox{0.6\linewidth}{!}{ 
%\begin{center}
\resizebox{0.3\linewidth}{!}
{
    \begin{tikzpicture}
    % \begin{axis}[%
    %   width=.8\linewidth,
    % %   height=6cm, 
    %     legend columns = 5,
    %     hide axis,
    %   ]
    \begin{axis}[%
        hide axis,
        xmin=10, xmax=50,
        ymin=0,ymax=0.4,
        legend columns=-1,
        legend style={draw=white!15!black,legend cell align=left}
        ]
        % \addlegendimage{empty legend}
        %\addlegendimage{}
        %\addlegendentry{}
        \addlegendimage{ttl1}
        \addlegendentry{$R_\mathcal{S}$}
        \addlegendimage{ttl2}
         \addlegendentry{FID}
         \addlegendimage{ttl4}
        \addlegendentry{Time conditioning}
        \addlegendimage{ttl3}
         \addlegendentry{Temperature changes}
        %\addlegendimage{empty legend}
        %\addlegendentry{$\times$: feature missing}
          %\addlegendentry{cat}
        %\addlegendimage{y3c}
        %\addlegendentry{cat,ar}
    %\addlegendentry{\texttt{cat \& ar}}
    \end{axis}
    \end{tikzpicture}
    }
%\end{center}
}\\
\begin{tabular}{lr}
\resizebox{0.5\linewidth}{!}{
\begin{tikzpicture}
\begin{axis}[%
%   width=.5\linewidth,
  height=6cm, 
   xlabel={{\Large Observed time steps}},
   %xlabel shift=-1.1ex,
%   xtick = {64, 128, 256, 512, 1024},
%   xticklabels = {64, 128, 256, 512, 1024},
  %ylabel={mAP}, 
  minor tick num=3,
  ylabel={\Large $R_\mathcal{S} (\uparrow)$},
%ylabel shift=-1.1ex,
%   ymin=0.595, ymax=0.645,
    % xmin=50, xmax=700,
  %legend size=small,
  legend pos=north east,
    % legend pos=outer north east,
    % xmode=log,
    %tick label style={font=tiny}
    yticklabel style = {font=\Large},
    xticklabel style = {font=\Large},
    title={{\Large Sample quality vs time conditioning}},
  ]
\pgfplotstableread{
iter    map acc rs fid
0  32.3   53.49  0.95 3.14

4  36.2   54.0   1.06 3.0
8  41.0   56.7   1.20 2.92
16 44.2   57.7   1.3  2.78

}{\map}
%2  34.4   54.2   1.01 3.26

\pgfplotstableread{
temp  tmap  trs  tacc tfid
0.8   37.0  1.08    57 3.9
0.9   34.7  1.05    55 3.4
0.95  33.2  0.97    54 3.26
1.0   32.3  0.95    43.5 3.14
1.05  31.3  0.92    52 2.99
1.1   29.9  0.88    51 2.94
1.2   27.0  0.79    49 2.63

}{\map}
%2  34.4   54.2   1.01 3.26
    \addplot[tt3,line width=2pt,mark size=3pt]      table[x=iter,  y=rs]   \map; 
    %\addplot[y2c]      table[x=iter,  y=cat]   \map; 
    %\addplot[y3c]      table[x=iter,  y=mlp_ar]   \map; 
\end{axis}
\begin{axis}[%
    %yticklabel style = {font=\footnotesize,xshift=-0.5ex},
    %xticklabel style = {font=\footnotesize,yshift=-0.5ex},
    %xlabel shift=-1.1ex,
    ylabel={\Large FID $(\downarrow)$},
    %ymin=2.5, ymax=4,
    %ylabel shift=-1.5ex,
    %xlabel={{Softmax temperature}},
    tick style={grid=none},
    ytick style={draw=none},
    yticklabel style = {font=\Large},
    xticklabel style = {font=\Large},
    minor tick num=3,
      height=6cm, 
      %scale only axis,
      %xmin=0,xmax=15,
      %domain=0:15,
      axis y line*=right]
      %axis x line*=top]
      %\addplot[red] {x};
%   \begin{axis}[
%         xmin=0, xmax=0.2, % x scale
%         ymin=0, ymax=1, % y scale
%         domain=0:1,  % added, key improvements
% ]
% \addplot    {1/(1e9*x/11.778*3.273e-8 - 3.273e-8 + 1)};
% \addplot    coordinates{(.125, .744)};
% \end{axis}

% \end{tikzpicture}
% \begin{tikzpicture}
% \begin{axis}[
%         xmin=0, xmax=1, % x scale
%         ymin=0, ymax=1, % y scale
%         domain=0:1   % added, key improvements
% ]
\pgfplotstableread{
iter    map acc rs fid
0  32.3   53.49  0.95 3.14

4  36.2   54.0   1.06 3.0
8  41.0   56.7   1.20 2.92
16 44.2   57.7   1.3  2.78

}{\map}
%2  34.4   54.2   1.01 3.26

\pgfplotstableread{
temp  tmap  trs  tacc tfid
0.8   37.0  1.08    57 3.9
0.9   34.7  1.05    55 3.4
0.95  33.2  0.97    54 3.26
1.0   32.3  0.95    43.5 3.14
1.05  31.3  0.92    52 2.99
1.1   29.9  0.88    51 2.94
1.2   27.0  0.79    49 2.63

}{\map}
%2  34.4   54.2   1.01 3.26
    \addplot[tt4,line width=2pt,mark size=3pt]      table[x=iter,  y=fid]   \map; 
    %\addplot[y2c]      table[x=iter,  y=cat]   \map; 
    %\addplot[y3c]      table[x=iter,  y=mlp_ar]   \map; 
\end{axis}
\end{tikzpicture}} & 
\resizebox{0.5\linewidth}{!}{
\begin{tikzpicture}
\begin{axis}[%
%   width=.5\linewidth,
  height=6cm, 
  xlabel={{\Large Softmax Temperature }},
   %xlabel shift=-1.1ex,
  %xtick = {64, 128, 256, 512, 1024},
  %xticklabels = {64, 128, 256, 512, 1024},
  %ylabel={mAP}, 
  minor tick num=3,
  ylabel={\Large $R_\mathcal{S} (\uparrow)$},
  %ylabel shift=-1.1ex,
%   ymin=0.595, ymax=0.645,
    % xmin=50, xmax=700,
  %legend size=small,
  legend pos=north east,
    % legend pos=outer north east,
    % xmode=log,
    %tick label style={font=tiny}
    yticklabel style = {font=\Large},
    xticklabel style = {font=\Large},
    title={{\Large Quality vs softmax temperature}},
  ]
\pgfplotstableread{
iter    map acc rs fid
0  32.3   53.49  0.95 3.14

4  36.2   54.0   1.06 3.0
8  41.0   56.7   1.20 2.92
16 44.2   57.7   1.3  2.78

}{\map}
%2  34.4   54.2   1.01 3.26

\pgfplotstableread{
temp  tmap  trs  tacc tfid
0.8   37.0  1.08    57 3.9
0.9   34.7  1.05    55 3.4
0.95  33.2  0.97    54 3.26
1.0   32.3  0.95    43.5 3.14
1.05  31.3  0.92    52 2.99
1.1   29.9  0.88    51 2.94
1.2   27.0  0.79    49 2.63

}{\map}
%2  34.4   54.2   1.01 3.26
    \addplot[tt1,line width=2pt,mark size=5pt]      table[x=temp,  y=trs]   \map; 
    %\addplot[y2c]      table[x=iter,  y=cat]   \map; 
    %\addplot[y3c]      table[x=iter,  y=mlp_ar]   \map; 
\end{axis}
\begin{axis}[%
%   width=.5\linewidth,
  height=6cm, 
%   xlabel={{\Large Softmax Temperature }},
   %xlabel shift=-1.1ex,
  %xtick = {64, 128, 256, 512, 1024},
  %xticklabels = {64, 128, 256, 512, 1024},
  %ylabel={mAP}, 
  minor tick num=3,
  ylabel={\Large FID $(\downarrow)$},
  tick style={grid=none},
  ytick style={draw=none},
  xtick style={draw=none},
  %ylabel shift=-1.1ex,
%   ymin=0.595, ymax=0.645,
    % xmin=50, xmax=700,
  %legend size=small,
  legend pos=north east,
  ymin=2, ymax=4.3,
    % legend pos=outer north east,
    % xmode=log,
    %tick label style={font=tiny}
    yticklabel style = {font=\Large},
    xticklabel style = {font=\Large},
     axis y line*=right]
\pgfplotstableread{
iter    map acc rs fid
0  32.3   53.49  0.95 3.14

4  36.2   54.0   1.06 3.0
8  41.0   56.7   1.20 2.92
16 44.2   57.7   1.3  2.78

}{\map}
%2  34.4   54.2   1.01 3.26

\pgfplotstableread{
temp  tmap  trs  tacc tfid
0.8   37.0  1.08    57 3.9
0.9   34.7  1.05    55 3.4
0.95  33.2  0.97    54 3.26
1.0   32.3  0.95    43.5 3.14
1.05  31.3  0.92    52 2.99
1.1   29.9  0.88    51 2.94
1.2   27.0  0.79    49 2.63

}{\map}
%2  34.4   54.2   1.01 3.26
    \addplot[tt2,line width=2pt,mark size=5pt]      table[x=temp,  y=tfid]   \map; 
    %\addplot[y2c]      table[x=iter,  y=cat]   \map; 
    %\addplot[y3c]      table[x=iter,  y=mlp_ar]   \map; 
\end{axis}
% \begin{axis}[%
%     %yticklabel style = {font=\footnotesize,xshift=-0.5ex},
%     %xticklabel style = {font=\footnotesize,yshift=-0.5ex},
%     %xlabel shift=-1.1ex,
%     ylabel={\Large FID $(\downarrow)$},
%     %ylabel shift=-1.5ex,
%     %xlabel={{Softmax temperature}},
%     tick style={grid=none},
%     ytick style={draw=none},
%     xtick style={draw=none},
%     minor tick num=3,
%     ymin=2.9, ymax=4.3,
%     height=6cm, 
%     yticklabel style = {font=\Large},
%     xticklabel style = {font=\Large},
%     %ymin=2, ymax=5,
%       %scale only axis,
%       %xmin=0,xmax=15,
%       %domain=0:15,
%       axis y line*=right]
%       axis x line*=top]
%       \addplot[red] {x};
%   \begin{axis}[
%         xmin=0, xmax=0.2, % x scale
%         ymin=0, ymax=1, % y scale
%         domain=0:1,  % added, key improvements
% ]
% \addplot    {1/(1e9*x/11.778*3.273e-8 - 3.273e-8 + 1)};
% \addplot    coordinates{(.125, .744)};
% \end{axis}

% % \end{tikzpicture}
% % \begin{tikzpicture}
% % \begin{axis}[
% %         xmin=0, xmax=1, % x scale
% %         ymin=0, ymax=1, % y scale
% %         domain=0:1   % added, key improvements
% % ]
% \input{plots/bottleneck/read_time_cond}
%     \addplot[tt2,line width=2pt,mark size=5pt]      table[x=temp,  y=tfid]   \map; 
%     %\addplot[y2c]      table[x=iter,  y=cat]   \map; 
%     %\addplot[y3c]      table[x=iter,  y=mlp_ar]   \map; 
% \end{axis}
\end{tikzpicture}}
\end{tabular}
\end{center}%\\[-0.6cm]
  \caption{\footnotesize \label{fig:time_cond}\footnotesize \textbf{Sample quality} with our best model for different amounts of observed motion, and different temperatures, measure with the FID and $R_\mathcal{S}$ metrics.}%
\end{figure}

\noindent\textbf{Ablations on next index prediction.}
We now study two design choices made in the GPT component of \ours: the choices of input embedding and predictions head.
Using the proper input embedding has a strong impact on the performance of transformer architectures~\cite{Radford2018ImprovingLU}, and depends on the input data.
In Table \ref{tab:abl_input}, we study this impact when working with human motion. 
For this ablation, we fix the latent sequence space, \ie, the auto-encoder hyper-parameters and weights, and train our GPT model with different input embeddings. 
We measure sample quality, and the accuracy of the GPT model at predicting discrete sequence indices. Note that this accuracy is directly comparable across models, as the latent space is frozen and identical.

In the first row, we observe that embedding the action at each timestep, rather than as an extra transformer input, has significant positive impact.
Conditioning on the sequence length is also beneficial (Row 3 \vs Row 4); this is expected, as it relieves the model from having to predict when to stop generating new poses. 
As an added benefit, it also allows extra control at inference time.
We also see that a concatenation of the embedded information, followed by a linear projection -- which can be seen as a learned weighted sum -- is better than simple summation on BABEL. On the other hand this extra model capacity is not beneficial on HumanAct12, which may be due to the size of the dataset. In Figure~\ref{fig:sumcat}, we also observe that models using concatenation rather than summation train significantly faster.

Having determined the best input configuration for both datasets, we further experiment with more expressive output layers for the model; we show that having a MLP head rather than a single fully-connected layer is beneficial, and we obtain further gains using an auto-regressive layer (see Section~\ref{sec:method:gpt}). This can be explained by the fact that with product quantization, several codebook indices are extracted simultaneously from a single input vector, but are not independent, and using an MLP and/or an auto-regressive layer better captures the correlations between them. \\

\begin{figure}[tb]
\begin{floatrow}
\capbtabbox[5cm]{%
\resizebox{0.7\linewidth}{!}{
    \begin{tabular}{cccc}
    \toprule
    \begin{tabular}{c}$K$ \\[-0.2cm] {\tiny (nb. codebooks)  }\end{tabular} &
	\begin{tabular}{c}{ \small Causal} \\ E  \end{tabular} &
	\begin{tabular}{c}{ \small Causal} \\ D  \end{tabular} &
	pve ($\downarrow$) \\
        \midrule                                                                          
	2 & $\times$ & $\times$ & \bf{35.7} \\
	2 & \checkmark & $\times$ & 38.3 \\
	2 & \checkmark & \checkmark & 50.2  \\
	\midrule
	4 & $\times$ & $\times$ & \bf{25.3}  \\
	4 & \checkmark & $\times$ & 28.2 \\
	4 & \checkmark & \checkmark & 38.6\\
	\bottomrule
    \end{tabular}}
}{%
  \caption{\small \label{tab:causal}\textbf{Impact of causal attention} in the encoder and decoder for $C=256$.}%
}
\ffigbox[6cm]{%
\begin{center}
\resizebox{1\linewidth}{!}{
%\hspace*{-0.5cm}
\begin{tikzpicture}
\begin{axis}[%
%   width=.5\linewidth,
  height=3.5cm, 
   xlabel={{\tiny Nb. iterations}},
  minor tick num=3,
%   ymin=0.595, ymax=0.645,
    % xmin=50, xmax=700,
  %legend size=small,
  legend pos=north east,
  xticklabels={$2$, $4$, $16$, $64$, $256$},
  xlabel shift=-1.1ex,
  xtick={1,2,4,6,8},
  yticklabel style = {font=\tiny,xshift=0.5ex},
  xticklabel style = {font=\tiny,yshift=0.5ex},
   % legend pos=outer north east,
    % xmode=log,
    %tick label style={font=tiny}
    title={{\tiny FID ($\downarrow$)
    }},
            title style={yshift=-8pt},
  ]
% \pgfplotstableread{
% repeat    PoseGPT 
% 0         2.80      
% 1         3.69     
% 2         3.43
% 4         3.57
% 8         3.49
% 16        3.47
% 32        3.60
% 64        3.57
% 128       3.54
% 256       3.61
% 512       3.58
% }
\pgfplotstableread{
iter logiter    PoseGPT 
1      0.0                          2.80      
2      1.0               	    3.69     
3      1.584962500721156 	    3.43
5      2.321928094887362 	    3.57
9      3.169925001442312 	    3.49
17     4.087462841250339 	    3.47
33     5.044394119358453 	    3.60
65     6.022367813028454 	    3.57
129    7.011227255423254 	    3.54
257    8.005624549193879 	    3.61
513    9.002815015607053 	    3.58

}
{\map}

    \addplot[r1c]      table[x=logiter,  y=PoseGPT]   \map; 
    % \addplot[y2c]      table[x=repeat,  y=MKT]   \map; 
    % \addplot[y3c]      table[x=repeat,  y=mlp_ar]   \map; 
\end{axis}
\pgfresetboundingbox
\path
  (current axis.south west) -- ++(0.0in,0.0in)
  rectangle (current axis.north east) -- ++(0.0in,0.0in);
\end{tikzpicture}}
\end{center}
}{%
  \caption{\label{fig:error_drift}\footnotesize \textbf{Evaluation of error drift. } Model iteratively conditioned on last predictions made.}%
}
\end{floatrow}
\end{figure}

\noindent\textbf{Causal attention.}
In Table~\ref{tab:causal}, we study the impact of using causal attention in the auto-encoder, for $K \in \{2, 4\}$ and $C = 256$. Causal attention is a restriction on model flexibility, as it limits the inputs used by features in the encoder. Empirically, we observe that adding causal attention indeed degrades performance.
 Adding it to the encoder, which is mandatory to create a model that can be conditioned on past observations, causes only a mild degradation. Adding it in the decoder as well allows to run the model on-line, \ie, make observations and predictions in parallel, but strongly degrades performance.\\

\noindent\textbf{Conditioning and temperature.} Conditioning the model on past observation is expected to improve the quality of generated samples. In Figure~\ref{fig:time_cond} (left), we see that indeed both $R_\mathcal{S}$ and the FID improve monotonically as the length of observations increases. In the right plot, we see that increasing or decreasing the softmax temperature leads to a trade-off between the two metrics; this behaviour can be expected: decreasing the temperature improves sample quality by concentrating the mass on major modes of the distribution, and thus increases mode-dropping.\\
\noindent\textbf{Error-drift in long-term horizon generation. }
In Figure \ref{fig:error_drift}, we study the robustness of \ours to error drift, a typical failure case of models that make auto-regressive predictions in continuous space.
To this end, we sample from our model several times consecutively, by conditioning the last pose generated by the model. To initiate this process, the first motion is generated without temporal conditioning.
Empirically, we observe that in this setting, \ours is robust to long-term error drift: the FID initially degrades but remains stable even when we repeat the generation process many times.

\subsection{Comparison to the state of the art}
\label{sec:expe:humanact12}

In Table \ref{tab:sota_hum_bab}, we compare \ours against the state-of-the-art results. 
For fair comparison, these metrics are computed without conditioning on past observations.
We find that \ours outperforms the state-of-the-art method, namely ACTOR \cite{actor}, when looking at the FID metric with a relative gain of $33\%$ ($0.12$ \vs $0.08$) on HumanAct12 and over $50\%$ on BABEL. 
The performance in diversity and the multimodality indicates that \ours covers the human motion distribution of this dataset.
On BABEL, the gains are around $50\%$ in terms of both FID and classification accuracy. The gains in classification accuracy indicate both higher quality samples, and a richer distribution. \\

\begin{table}[!b]
    %\\[-0.3cm]
    \centering
    \begin{tabular}{ccc}
    \resizebox{.39\textwidth}{!}{
    \begin{tabular}{lcccc}
    \toprule
        Model &  FID$\downarrow$ & $R_\mathcal{T}$ (\%).$\uparrow$ & Div. & Multimod. \\
        \midrule 
        Real & 0.02 & 99.4 & 6.86 & 2.60 \\
        Action2Motion \cite{chuan2020action2motion} & 2.46 & 92.3 & 7.03 & 2.87 \\
        ACTOR \cite{actor} & 0.12 & 95.5 & 6.84 & 2.53 \\
        \textbf{\ours} &  \textbf{0.08} & \textbf{95.8} & 6.85 & 2.82 \\
    \bottomrule
    \end{tabular}
    } & 
    \resizebox{.36\textwidth}{!}{
    \begin{tabular}{lHHcccr}
    \toprule
        Model & action cond. & length cond. & future pred. & $R_\mathcal{S} (\uparrow)$    & FID$\downarrow$  & $R_\mathcal{T} (\uparrow)$ \\
        \midrule
        Real & - & - & - & 1.0 & 0.01 & 1.0 \\
	\midrule
        ACTOR$^*$ & \checkmark & \checkmark & X & 0.35 & 9.5 & 0.56\\
        PoseGPT & \checkmark & \checkmark & \checkmark & \bf{0.64}  & \bf{2.7} & \bf{0.98}\\
    \bottomrule
    \end{tabular}} & 
    \resizebox{.22\textwidth}{!}{
    \renewcommand{\arraystretch}{1.3}
    \begin{tabular}{cccH}
    \toprule
        Model &  FID $\downarrow$ & $\text{Acc}_S(D)$ $\uparrow$ &	$\text{Acc}_D(S)$ $\uparrow$ \\
        \midrule    
        Real & 0.01 & 0.99 & - \\
        \midrule 
        ACTOR* & 20.7 & 0.20 & - \\
        \ours & \bf{5.1} & \bf{0.86} & - \\
        \bottomrule
    \end{tabular}
    }
    \end{tabular}\\[-0.1cm]
    \caption{\label{tab:sota_hum_bab}\textbf{State-of-the-art comparison.}  On HumanAct12 (left), \ours obtains better FID and comparable classification accuracy. On BABEL (center) and on GRAB (right), \ours obtains substantial gains for all metrics.
    $^*$ means trained by us based on official code. Note that the FID of real data is not $0$ due to data augmentations.
    For consistency with \cite{actor,chuan2020action2motion}, we report diversity and multimodality metrics on HumanAct12; these metrics are considered good when close to the values obtained on real data.
}
\end{table}

\begin{figure}[t!]
\centering
\includegraphics[width=0.98\linewidth]{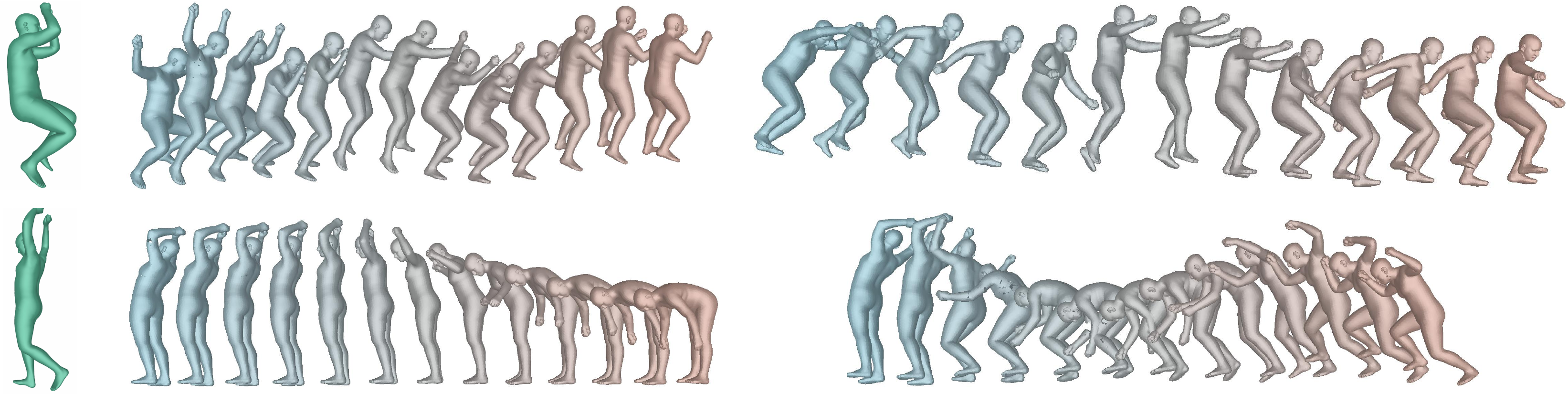}\\[-0.1cm]
\caption{\label{fig:samples_2}\textbf{Samples conditioned on past observation.} 
On the left in green, we show an observed initial pose, then we sample two different future human motions that we show side by side.
The top row corresponds to the human action `jumping' and the bottom row is sampled from the human action `stretching'.
}
\end{figure}

\begin{figure}[t!]
\centering
\includegraphics[width=0.98\linewidth]{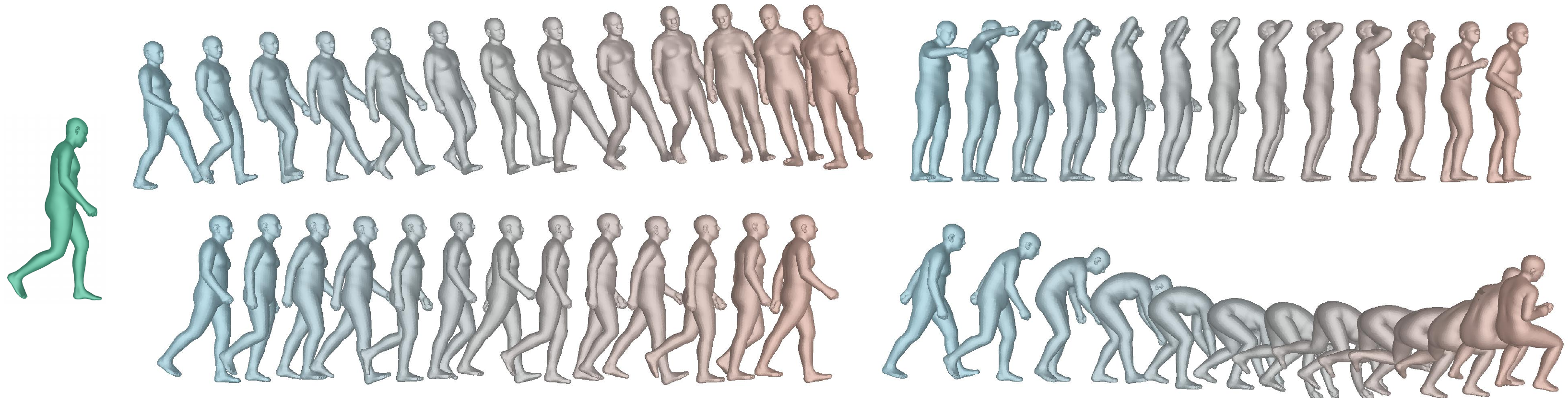}\\[-0.1cm]
\caption{\label{fig:samples_3}\textbf{Samples conditioned on an initial pose and with four different actions.}
	Given an initial pose shown in green, we generate four different human motions conditioned on four different actions. What are these actions?$^{\ref{myfootnote}}$
}
\end{figure}

\noindent\textbf{Qualitative examples. }
Finally, we show samples of human motions generated by \ours.
In Figure~\ref{fig:samples_1}, we show samples of human motion generated by conditioning on a human action only.
We observe that human motions are realistic and diverse for both actions.
Then in Figure~\ref{fig:samples_2}, we display two possible future motions given an initial pose and an action.
The generated human motions are diverse which demonstrates that \ours is able to handle the multimodal nature of the future.
Finally in Figure~\ref{fig:samples_3}, given a initial pose, we generate four human motions with four different actions\footnote{\label{myfootnote}\raisebox{\depth}{\scalebox{-1}[-1]{From left to right and top to bottom: `turning', `touching face', `walking', `sitting'.}}}.
This demonstrates that the action information is taken into account and impacts the human motion generation.
We provide more visualizations in the supplementary material.

\section{Conclusion}
This work introduces \ours, an auto-regressive transformer-based approach which quantizes human motion into latent sequences.
Given a human action, a duration and an arbitrarily long past observation, it outputs realistic and diverse 3D human motions.
We provide quantitative and qualitative experiments to show the strengths of our proposed method.
In particular, ablations demonstrate that quantization is a key component, and we study each part of our approach in detail.
\ours reaches state-of-the-art performance on three different benchmarks and is able to generate human motions given an action label, conditioned on observed past motion of arbitrary length.%, and is the first model to scale to the large-scale BABEL dataset.
\newpage
{
\small
\bibliographystyle{splncs04}
\bibliography{ref}
}
\newpage

%We now present additional details about \ours (Section~\ref{app:info}). %We will release an implementation of our work upon paper acceptance to ease reproduction of our work. 

% \section{Discussion on the attached video}
% \label{app:video}

% \noindent\textbf{Samples generated without observed motion.} 
% First we show some samples generated from scratch for four different human actions, namely `walk', `turn', `jump' and `dance'.
% The initial human poses are different for all samples and the human motions are diverse; this indicates that \ours is able to generate realistic, diverse and discriminative human motions.
% We also observe that even though the human action `dance' is not well represented in the BABEL dataset, \ours is still able to generate diverse samples. \\

% \noindent\textbf{Samples conditioned on an initial human pose.} 
% We then show samples obtained while conditioning our model on a initial pose for the human actions `run' and `turn'.
% There is high diversity in the future human motions generated by \ours, yet all samples are realistic futures given the initial human pose, which demonstrates the flexibility of the model. \\

% \noindent\textbf{Samples conditioned on an observed human motion.}
% Finally we generate samples while conditioning the model on 10 frames, and visualize future motions for the classes `turn', `throw' and `stretch'.
% We observe that increasing the duration of the observations allows us to decrease the uncertainty on the predicted futures, even though they remain diverse and do not mode-drop to a single modality.

%%%% Supplementary
\appendix

\section{Appendix}
\label{app:info}

\noindent\textbf{Auto-regressive prediction head.} 
In Section \textcolor{red}{3.1} of the main paper, we propose to model correlations between the $K$ codebook indices produced by product quantization by using a prediction head that is auto-regressive over the $K$ codebooks.
At train time, this prediction head takes as input (i) the logits produced by the standard head of the network and (ii) the $K$ indices embedded using the same embedding as used for the inputs.
Then index $k$ is predicted from a concatenation of (\texttt{logits}$_{1 \hdots k-1}$, \texttt{embed(Input)}$_{k \hdots K}$).
Note that this induces almost no overhead: $K-1$ extra linear layers, used in parallel, with $K$ typically in $\{2, 4\}$. 
This stands in contrast with the naive solution which would be to concatenate the products along the time dimension: it would have increased the cost of the whole network by a factor $K^2$ due to the quadratic cost of self-attention mechanisms.
At sample time, the $K$ indices predicted at a time step $t$ are sampled sequentially and used as input for the next prediction; thus, the $k-th$ token is predicted conditionally on tokens $0\hdots k-1$ as is the case at train time.
Importantly, this only requires running the \emph{prediction heads} sequentially rather than in parallel as is done at train time.
On the other hand, the naive solution would have increased the sampling time by a factor $K$. \\

\noindent\textbf{Input embedding.}
Self-attention based architectures are invariant to permutations of their input, which avoids inductive biases in the architecture.
However this can be detrimental when modeling sequential data, as positional information is lost; we follow the standard remedy of learning 1-D positional encodings, added to the embedded input data, to account for the temporal dimension.
We also learn embeddings for each action token $a$ and for the sequence length $T$. 
We experiment with two ways of adding this conditioning information to the input data: the first one is by adding an extra token to the input sequence, which is possible because the $a$ and $T$ are constant across time.
The second is to add the information at every time step. For this we again test two strategies: embedding all informations separately and simply summing them, or concatenating the embeddings and learning an extra layer on top of that.
More precisely, each embedding --input, positional, action and length -- is a $D_{\text{emb}}-$dimensional vector; they are concatenated together, linearly projected again to a $D_{\text{emb}}-$ dimensional space and fed to the transformer model.  
We find in the experimental section (Section \textcolor{red}{4.2}) that the last strategy is the best one. \\

\noindent\textbf{Architectural details.}
All three components - $E$, $D$, and $G$ are based on transformers. The encoder $E$ is a stack of 3 blocks where each block is composed of 4 transformer layers.
We perform temporal downsampling by a factor of 2 after the first block by default, and after each following block when downsampling by a factor greater than two (i.e., when $T/T_d \in \{4,8\}$).
The input embedding dimension of each block is 512.
each transformer layer is composed of a self-attention module which has 5 attention heads -- each of dimension 32 -- and a feedforwad layer with 512 hidden units.
We use a dropout of 0.1  both in the self-attention and in feed-forward modules.
The decoder $D$ mirrors the decoder, with the same hyper-parameters and blocks called in reversed order.
The auto-regressive model $G$ is a transformer which has 8 layers; the self attention blocks use 4 attentions heads, each of dimension 256. The input embedding dimension is 256, and we use a dropout of 0.2 for this network.\\

\noindent\textbf{Training details.} 
We implement \ours in Python using the PyTorch framework \cite{paszke2019pytorch} and we train our network from scratch using the Adam optimizer \cite{kingma2014adam} with a learning rate of $5.10^{-5}$ and default parameters.
Both the auto-encoder in the first training stage and the auto-regressive network in the second stage are trained for 2 million iterations.
The whole training procedure takes 3 days on one single Nvidia V100 GPU. $L_2$ reconstruction losses are applied to body model parameters as well as directly on vertices for a randomly sampled subset of time steps for efficiency/memory reason; we found that using $10\%$ to $20\%$ of the frames is sufficient.\\

\noindent\textbf{Action classifiers.} 
For HumanAct12 we use the classifier provided by \cite{chuan2020action2motion} which is a GRU followed by a a fully-connected layer.
The classifier takes as input 3D joints of the human skeleton centered around the spine.
For both BABEL and GRAB, we train the classifier ourselves using the same architecture as described above.
\\

% \newpage
% {\small
% \bibliographystyle{splncs04}
% \bibliography{ref}}
\end{document}